%% file: main.tex
\documentclass[12pt]{iopart}
\usepackage{graphicx}
\usepackage{svg}
\expandafter\let\csname equation*\endcsname\relax
\expandafter\let\csname endequation*\endcsname\relax
\usepackage{amsmath}
\usepackage{caption}

\usepackage{hyperref}

\begin{document}

\title[The Robot of Theseus]{The Robot of Theseus: A modular robotic testbed for legged locomotion}

\author{Karthik Urs*,
Jessica Carlson*,
Aditya Srinivas Manohar,
Michael Rakowiecki,
Abdulhadi Alkayyali,
John E. Saunders,
Faris Tulbah,
Talia Y. Moore
}

\address{* Denotes co-first authorship\\
University of Michigan,
Ann Arbor, MI 48103}
\ead{taliaym@umich.edu}
\vspace{10pt}

\begin{abstract}
Robotic models are useful for independently varying specific features, but most quadrupedal robots differ so greatly from animal morphologies that they have minimal biomechanical relevance. 
Commercially available quadrupedal robots are also prohibitively expensive for biological research programs and difficult to customize.
Here, we present a low-cost quadrupedal robot with modular legs that can match a wide range of animal morphologies for biomechanical hypothesis testing. 
The Robot Of Theseus (TROT) costs $\approx$\$4000 to build out of 3D printed parts and standard off-the-shelf supplies. 
Each limb consists of 2 or 3 rigid links; the proximal joint can be rotated to become a knee or elbow.
Telescoping mechanisms vary the length of each limb link. 
The open-source software accommodates user-defined gaits and morphology changes. 
Effective leg length, or crouch, is determined by the four-bar linkage actuating each joint. 
The backdrivable motors can vary virtual spring stiffness and range of motion. 
Full descriptions of the TROT hardware and software are freely available online. 
We demonstrate the use of TROT to compare locomotion among extant, extinct, and theoretical morphologies.
In addition to biomechanical hypothesis testing, we envision a variety of different applications for this low-cost, modular, legged robotic platform, including developing novel control strategies, clearing land mines, or remote exploration. 
All CAD and code is available for download on the \href{www.embirlab.com/trot}{TROT project page}.
\end{abstract}

%
\vspace{2pc}

%
%
%

\input{sections/newintro.tex}
\input{sections/2design}

\input{sections/3experiments}

\input{sections/4results}
\input{sections/newdiscussion}

\section*{Acknowledgements}
The authors would like to acknowledge Mirielle Wong, Jack Withers, and Trinh Huynh for writing the assembly and user guides for TROT.
We would also like to thank the UM Robotics Walk lab for use of the treadmill and the ROAHM lab for use of the Phase Space motion capture cameras.
We would like to thank David Polly for sharing his table of limb measurements from the P.P. Gambaryan book.

\section*{Conflict of Interest}
The authors declare no conflicts of interest.

\section*{Funding}
The work was carried out with no external funding.

\section*{Data Availability Statement}
The data that support the findings of this study are available on the project page \href{www.embirlab.com/trot}.

\section*{References}
\bibliographystyle{ieeetr}
\bibliography{papers}

\end{document}

%% file: sections/newintro.tex
\section{Introduction}
\label{sec:intro}
Robotic physical models are powerful tools for understanding the function of biological structures.
Experimentally testing how changes in components of the robotic model affect overall function is an approach that helps us establish the mechanics of complex systems \cite{steinhardt2021physical,flyingfish2022,long2011testing}, understand how selective pressures have shaped the evolution of biomechanical phenotypes \cite{matthews2023genes,campos2017plant,long2006biomimetic}, and informs the design of engineered systems \cite{steinhardt2021physical,shah2021shape,long2011inspired}.
For example, the Salamandra Robotica demonstrates how a nervous system can control coordination of limb and body movement for both aquatic and terrestrial locomotion \cite{karakasiliotis2013we}.
Studies with robotic fish reveal how fin shape, size, location, and stiffness determine propulsion abilities \cite{lauder2007fish}.
In comparison to simulations, robotic models are especially useful because the validity of their results is independent of assumptions regarding how to model various physics phenomena \cite{long2007biomimetic}.

In the field of quadrupedal locomotion, robotic physical models span a wide range of bio-fidelity, from species-specific prototypes to generic body forms.
For example, a robot constructed by 3D printing CT scans of a fossil stem amniote revealed potential locomotion patterns for that species \cite{nyakatura2019reverse}.
This was a custom-created robot, requiring a large investment of time and resources to investigate a single species.
On the other end of the spectrum, commercially available robots have a generalized body form and are widely used as testbeds to explore the control of quadrupedal locomotion \cite{ANYmal2016,minicheetah2019,shi2018bio}.
These robots have demonstrated an incredible array of gaits and behaviors, but when there are significant morphological differences between robots and animals, the extent to which the robotic conclusions can transfer to biological insights is limited \cite{fukuhara2022comparative,ijspeert2014biorobotics}.
Few robots exist in the middle of the spectrum, where they would be useful for both biology and robotics studies.

Interdisciplinary research hints at the potential benefit of using robotic physical models to understand how changes in both morphology and control affect whole body dynamics.
For example, biomechanics researchers seeking to empirically test whether limb moment of inertia increases cost of transport in animals compared cheetahs and goats \cite{taylor1974running}. 
Although this relationship is intuitive, there were no significant differences in cost of transport between these two species, despite significant differences in limb moment of inertia. 
This unexpected result is potentially due to additional differences in kinematics, elastic energy storage, or other differences that co-evolved with limb moment of inertia in each species.
Similarly, the role of limb joint direction on whole body dynamics is informative when designing robots \cite{Smit2017,Raw2019}. 
These simulations suggest that there are significant whole-body tradeoffs associated with different limb configurations. 
However, because all simulations involve simplifying assumptions regarding contact physics \cite{tracy2023efficient}, it is unclear whether these conclusions hold true in the real world.
These are just some examples of biological and robotics questions that would benefit from empirically testing physical models that are sampled from a broad parameter space. 
Such an approach would require robotic models that can be easily varied in either morphology, control, or both.

We present The Robot of Theseus (TROT), a shape-shifting robotic physical model that was designed for testing both biological and robotic hypotheses. 
By altering either the morphology or the control of the robot, the onboard sensors can reveal how these changes, whether the result of evolution or engineering, affect whole-body locomotion dynamics. 
We describe the modular design of the robot, assess onboard sensor accuracy, and provide an example of a biological hypothesis that can be tested using this robot. 
We hope that TROT can facilitate more collaborative and mutually informative research between biology and robotics to advance our understanding and design of legged locomotion.

%% file: sections/2design.tex
\section{Robot Design}\label{sec:design}

TROT is a 10~kg modular quadrupedal robot designed to serve as a bridge between biomechanics and advanced robot control. 
Thus, we prioritized not only performance, but also reproducibility and accessibility for researchers without an engineering background. 
In this pursuit, the robot is equipped with high-performance actuation technology suitable for state-of-the-art control algorithms, while being constructed of only 3D printed and inexpensive off-the-shelf parts. 
In total, the robot can be built for less than \$4,000 USD.
Construction only requires the use of common hand tools and 3D printing (fused deposition modeling, FDM, and stereolithography, SLA). 
By leveraging additive manufacturing and off-the-shelf electronics, TROT may be built and modified in-house by most research labs. 

Our initial research application of TROT is to investigate the effect of limb morphology on whole body dynamics during locomotion. 
As such, the four limbs of the robot can be independently altered to form several morphological configurations and arbitrary link-to-link proportions with minimal reassembly. 
We also designed simple interfaces to easily integrate new components, such as different foot designs or additional sensors, if desired. 
All part files are posted on the ``The Robot of Theseus (TROT) CAD models and print files'' repository on the University of Michigan Deep Blue Data website \url{https://doi.org/10.7302/36xg-ez67}.

\subsection{Actuator Design}\label{sec:actuatordesign}

The limbs are each powered by three quasi-direct drive actuators, also known as proprioceptive actuators. 
These actuators are discussed in depth in a separate publication \cite{urs2022design}, but are summarized here.

Based on the actuation principles shared by several successful quadrupedal robots \cite{ANYmal2016} \cite{minicheetah2019} \cite{spot}, we designed the actuators to provide high torque outputs while maintaining very low output mechanical impedance (OMI) \cite{urs2022design}. 
This design philosophy enables predictable torque output and measurement via current control.
We used gap radius, time constant, and torque-specific inertia to select motors \cite{urs2022alternative} combined with an easily back-drivable planetary transmission (ratio 7.5:1 or 15:1). 
The actuators are driven by open-source, off-the-shelf motor drivers (Moteus r4.5/r4.8, mjbots, Cambridge, MA) that internally run positional and field-oriented current control at 40~kHz and expose a command and data interface for the motors over a data bus.



\subsection{Limb Design}\label{sec:legdesign}
The limbs are serially actuated rigid-body articulated mechanisms with a configuration-dependent number of links. 
In general, the limb has been designed to position the actuators proximally to the body to minimize limb inertia. 
To achieve this, we used linkages to transmit power distally.

\begin{figure}[ht]
  \centering
\includegraphics[width=\textwidth]{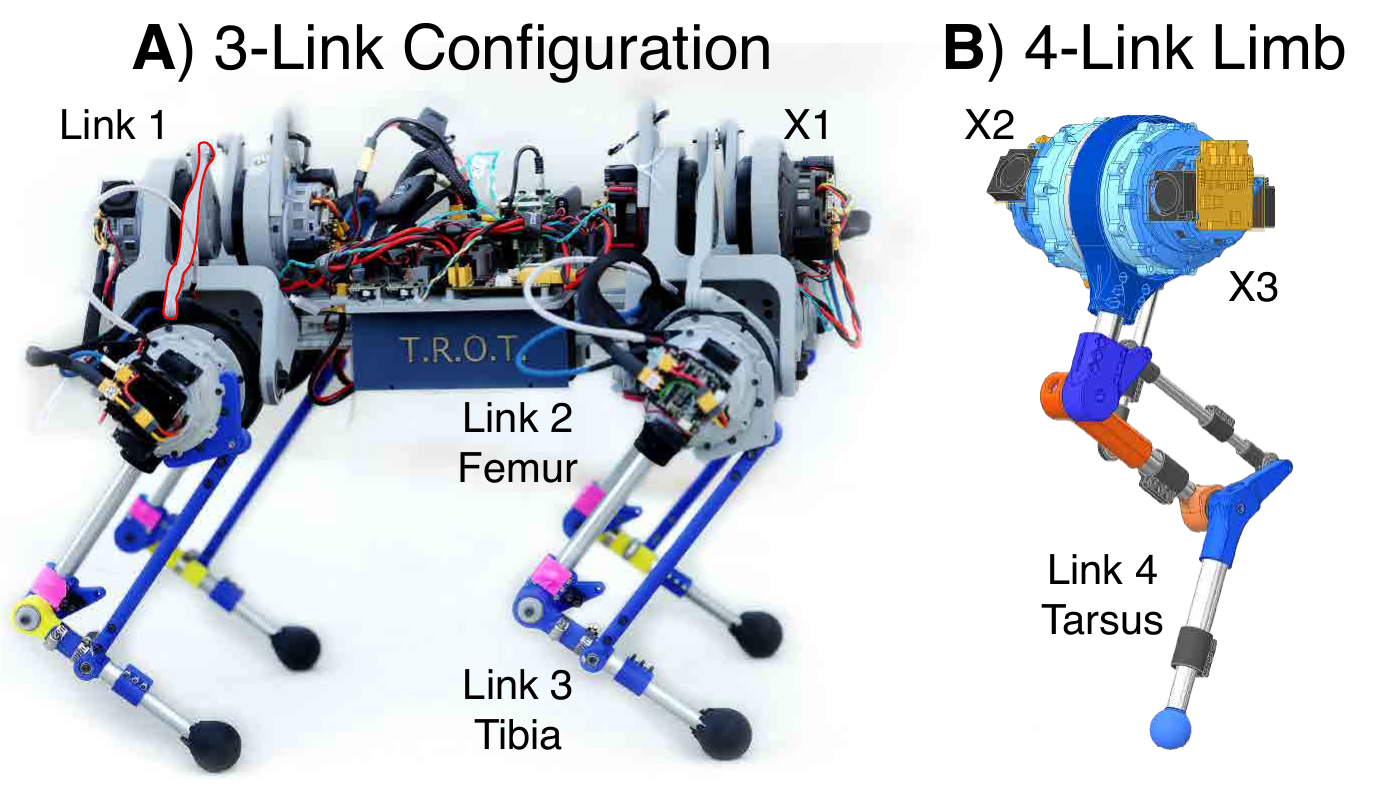}
\caption{Description of TROT features.
A) TROT in a three-link limb configuration with femur and tibia.
Link 1 is outlined in red.
B) In a four-link limb, the tarsus angle is parallel to the femur angle.
Limbs can be oriented with knees backward (as shown) or forward.}
\label{fig:trot_cad}
\end{figure}

Each limb interfaces with the TROT torso using a trunnion structure that allows the X1 actuator to move the limb medially and laterally via a linkage (link 1).
The linkage is connected to actuator X2, which swings the femur (link 2) in the fore-aft direction.
The output of X2 is connected to the X3 actuator, which controls the fore-aft motion of the tibia (link 3) with respect to femur.
The linkages attached to actuators X2 and X3 manipulate the limb to move anywhere in the sagittal plane. 
In the four-link configuration, the angle of the tarsus (link 4) is coupled to be parallel to the femur (link 2) through a 4-bar linkage, a simplification inspired by mammalian hindlimb kinematics \cite{fischer2002basic}.
The tarsus and tibia links are equipped with a telescoping mechanism allowing the user to extend the links to achieve the desired limb morphology, which is then locked in place with circumferential clamping.

\begin{figure}[ht]
  \centering
\includegraphics[width=\textwidth]{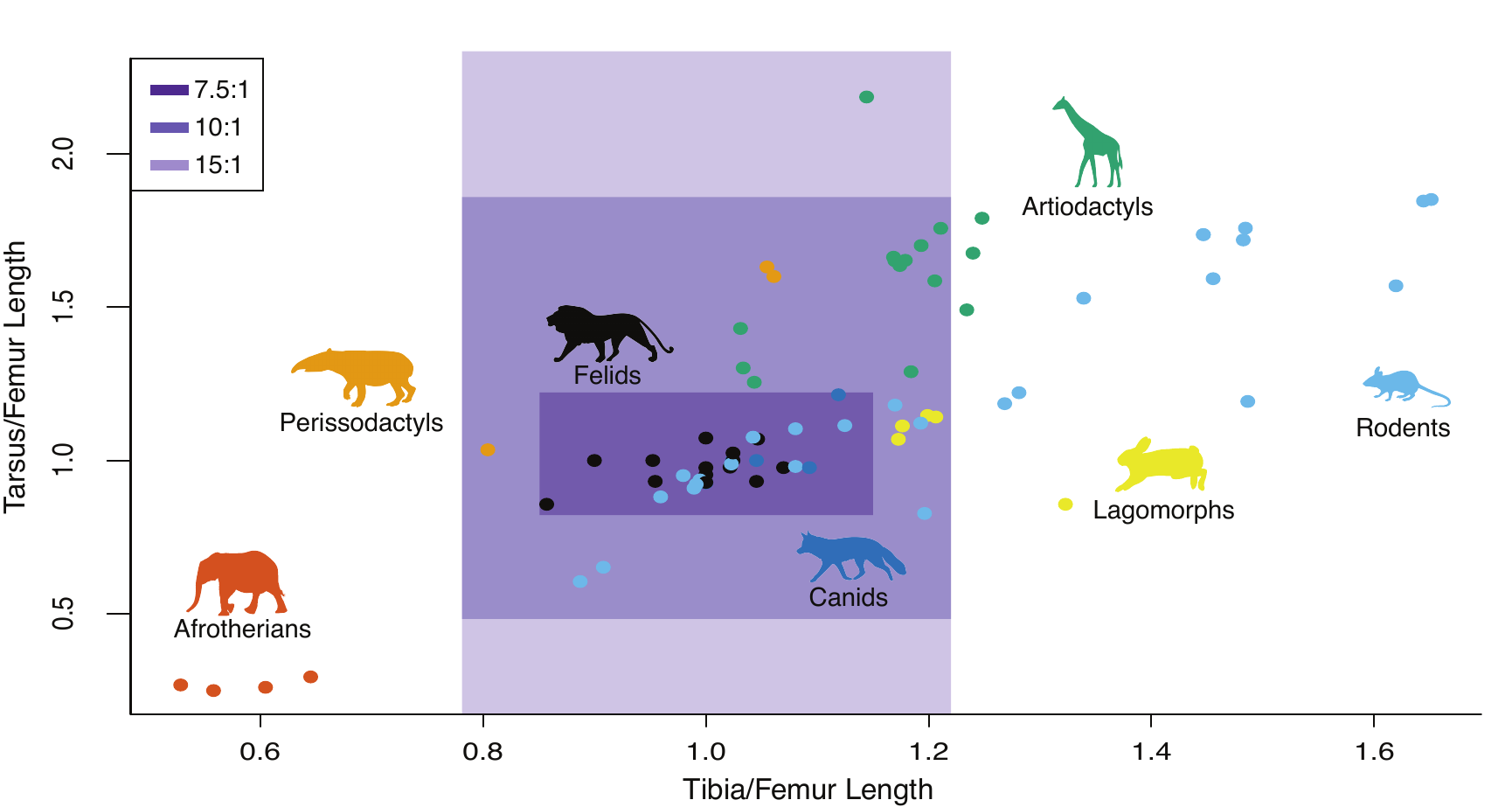}
\caption{A comparison of possible TROT limb length ratios (purple boxes) to animal limb length ratios across multiple families of quadrupedal mammals \cite{Gambaryan}.
The different shades of purple indicate the possible limb length ratios when constructed with gearboxes of different gear ratios.}
\label{fig:gear_ratio}
\end{figure}

Changing the limb length changes the moment of inertia of each limb segment, which therefore affects the torque required to move the limb segment.
Therefore, the maximum length for each segment is dictated by the maximum torque output of the motor and the gear ratio of the actuator (Fig. \ref{fig:gear_ratio})\cite{urs2022design}.
We used the equations of motion to simulate the maximum torques required to move each joint to find the range of limb lengths feasible with our existing actuators.
With a gear ratio of 7.5:1, TROT can have leg lengths corresponding to most felids, canids, and rodents.
Increasing the gear ratio from 7.5:1 to 15:1 greatly expands the range of limb lengths TROT can assume, which allows it to additionally emulate lagomorphs, artiodactyls, and perissodactyls.
If emulating Afrotherians is desired, TROT would require a different actuator.

\subsection{Torso Design}\label{sec:torsodesign}
The four limbs are rigidly mounted to a torso assembly. 
The primary structure is a segment of 40~mm T-slotted aluminum extrusion (80/20, Columbia City, IN), providing a rigid spine. 
The spacing between fore- and hindlimb actuators can be easily adjusted on the spine to vary the torso length.
Centered between the fore- and hindlimbs lie two 3D-printed cases that hold lithium polymer batteries, which serve as the power source for the whole robot. 
Above the cases is the electrical panel, a 3D-printed pegboard to which power circuitry and computers are mounted. 
The pegboard structure allows for easy reconfiguration of the panel to accommodate additional computation or sensing.

\subsection{Sensing}

The quasi-direct drive actuators can be used as sensors to reduce overall system complexity.
We used modern motor commutation to convert the current values into torque values \cite{mevey2009sensorless}, which can be directly used for many control strategies and state estimation techniques.
For example, the minimal elasticity of materials in the limb allows the motors in the quasi-direct drive actuators to act as sensors to measure ground-reaction forces (detailed in Sec. \ref{sec:exp:torquemeasurement}). 
To account for the lack of elastic elements in the limb, we vary the virtual spring stiffness of the limb by controlling the back drivability of the motors. 

We attached an Inertial Measurement Unit (IMU, ICM-42688, TDK, Tokyo, Japan) to the torso and encoders (AS5047P, AMS, Premstaetten, Austria) to each motor to collect position data.

\subsection{Computation and Control}\label{sec:computation}
TROT can operate in either a tethered or an untethered state.
Two onboard 15~V (``4S’’), 3.3~Ah lithium polymer batteries connected in series power TROT, for a main bus voltage of 30~V (nominally). 
The actuators are powered by this main bus, but smaller 5~V buses are created via step-down converters to power all other components.

The robot carries two single-board computers (Raspberry Pi 4B, Broadcom, Palo Alto, CA). 
The Robot Server Pi is dedicated to handling low-level operation, and the Robot Control Pi handles algorithmic control of the robot. 
They can communicate with each other via an ethernet connection through an onboard network switch (GS305, Netgear, San Jose, CA, USA). 

The Robot Server Pi was configured with CPU isolation and scheduler priority to run a custom C++ program in real time. 
This program (1) handles high-frequency communication with the actuators and IMU via an expansion board to interface with the data bus (pi3hat, mjbots, Cambridge, MA), (2) handles faults and maintains safe operating conditions, (3) exposes a simple command and query interface through Lightweight Communication and Marshaling (LCM) \cite{lcm2010}, and (4) logs data to an SSD. 
The Robot Control Pi leverages the open-source C++ `Cheetah-Software` for legged locomotion simulation and control that has been adapted to use LCM to interface with the robot \cite{minicheetah2019}. 
Using LCM and a network interface for high-level communication allows users to add an ethernet tether, wireless adapter, or more computers as necessary to provide monitoring or additional computational power.

The software builds on the opensource MIT Mini Cheetah controller \cite{cheetahgithub}.
We added functionality to accommodate TROT's unique morphological modularity by leveraging parameterized control in the existing code base to accommodate any changes to the robot’s morphology.
The `Cheetah-Software` package implements a quadratic program (QP) based model-predictive controller (MPC) with an optional whole-body controller (WBC) layer that, overall, generates actuator-level position, velocity, and/or torque commands. 


\subsection{Forward Kinematics}
\label{subsec:kinematics}

We developed a URDF to digitally represent the robot kinematics.
The linkage connections from X1 and X3 actuators technically form closed kinematic chains, but the tree-based approach to defining limb kinematics does not support closed-loop 4-bar linkages.
Therefore, to simplify computation of the contact Jacobian, we modeled the 3 link leg as though both actuators can be considered as acting at the link-to-link pivots.
This simplification greatly increases the speed of computation and has been previously demonstrated on another robot using 4-bar linkages in its legs \cite{gu2022overconstrained}.
This simplification results in a linear scaling error in the ground-reaction force computations, which can be manually calibrated for each axis.
The URDF and contact Jacobian files are available for download on the \href{www.embirlab.com/trot}{TROT project page}.

%% file: sections/3experiments.tex
\section{Experiments}\label{sec:experiments}
Here, we describe a set of experiments to demonstrate the capabilities of TROT and potential applications for biomechanical research.
All experiments were performed with TROT in a 3-link configuration, i.e., femur and tibia without tarsus.

\subsection{Onboard Sensor Validation}\label{sec:exp:torquemeasurement}
We manually controlled TROT to perform torso rotations in the roll, pitch, and yaw directions, while maintaining the same center of mass.  
During these rotations, we recorded ground reaction forces from a 6-axis force platform (BMS600600, AMTI, Watertown, MA).
We placed TROT on an L-shaped wooden platform so that only one limb was in direct contact with the force platform and all limbs were level.
We simultaneously recorded data from the TROT limb actuators and the force platform at 500~Hz.

Following the labeling system used in Figure \ref{fig:trot_cad}, we derived the Jacobian matrix for the end-effector of TROT's limb, $J(\theta)$: 

\begin{equation}
J_{foot}(\theta) = 
\begin{bmatrix}
   0 & L_{3}c_{23}+L_{2}c_{2} & L_{3}c_{23} \\
   L_{3}c_{1}c_{23}+L_{2}c_{1}c_{2}-(L_{1}+L_{4})s_{1}  & -L_{3}s_{1}s_{23}-L_{2}s_{1}s_{2} & -L_{3}s_{1}s_{23}\\
   L_{3}s_{1}c_{23}+L_{2}c_{2}s_{1}+(L_{1}+L_{4})c_{1} & L_{3}c_{1}s_{23}+L_{2}c_{1}s_{2} & L_{3}c_{1}s_{23}
   \end{bmatrix}
\label{eq:jacobian}
\end{equation}

with $c_{23} = c_{2}c_{3}-s_{2}s_{3}$, $s_{23} = s_{2}c_{3}+c_{2}s_{3}$, 
and $c_n$ equivalent to $cos(\theta{1})$.


Using the recorded torques $\tau$ in each of the limb's three actuators, and the Jacobian, $J(\theta)$ Eq.~\ref{eq:jacobian}, we calculated the ground reaction forces, GRF, for each limb using Eq.~\ref{eq:GRF}.

\begin{equation}
GRF = (J_{foot}(\theta)^{T})^{-1} \times \tau
\label{eq:GRF}
\end{equation}




\subsection{Effect of limb configuration on limb kinematics}\label{sec:exp:kinematicsetup}


The objective of this experiment was to demonstrate how altering limb length ratios would affect limb trajectories during locomotion while keeping the robot controller constant. 
We tested three limbs configurations, each totaling 400~cm, with tibia:femur ratios of 45:55, 50:50, and 55:45. 

For each unique limb ratio, we ran TROT in a trotting gait (contralateral legs in synchrony) at a constant speed of 0.1~m/s for one minute on a custom TuffTread treadmill (Conroe, Texas).
We used a Gamepad controller (F310, Logitech, Lausanne, Switzerland) to manually maintain a linear trajectory.
We recorded limb motions at 550~FPS with a Fastec IL-Series High-Speed Digital Camera placed perpendicular to TROT.
We digitized reflective markers on the limb to obtain kinematic data using the DLTdv8 package for MATLAB (R2022a, Natick, MA) \cite{hedrick2008software}.

\subsection{Effect of Limb Moment of Inertia on Cost of Transport}\label{sec:legconfig}

To demonstrate how TROT can be used to test biological hypotheses, we sought to empirically determine the effect of limb moment of inertia on Cost of Transport (COT), which has previously been difficult to isolate from other physiological variables when comparing different species \cite{taylor1974running}.

For this experiment, we used TROT in the 50:50 limb ratio configuration for all trials.
We added 50~g to either to all the femora (low moment of inertia configuration) or all the tibiae (high moment of inertia configuration).
We measured the moment of inertia of each limb empirically using the pendulum method \cite{dowling2006uncertainty}. 

We then ran TROT in a straight line using a trotting gait for three meters at a speed of 0.3~m/s. 
We analyzed 1000 samples of data from each experiment.

To find the cost of transport for each configuration, we first computed the current, $I$, by multiplying the torque applied to the actuators, $\tau$, by the actuator's torque constant, $K_{\tau}$, empirically determined in a previous publication \cite{urs2022design}. 

\begin{equation}
I = \tau K_{\tau}
\label{eq:current}
\end{equation}

Next, we used current data to compute the total electrical power, $P$, delivered to each actuator using Eqs.~\ref{eq:power} and \ref{eq:voltage}:


\begin{equation}
    P = IV
\label{eq:power}
\end{equation}
\begin{equation}
    V = \omega K_B * IR
    \label{eq:voltage}
\end{equation}
in which the product of the motor angular velocity, $\omega$, multiplied by the back-EMF constant, $K_E$, represents the back EMF voltage and the product of current, $I$, and phase resistance, $R$, represents the voltage drop across the motor winding.
The back-EMF constant was empirically determined in a previous publication \cite{urs2022alternative}.
We only counted positive values of power draw.

\begin{figure}[t]
  \centering
\includegraphics[width=\textwidth]{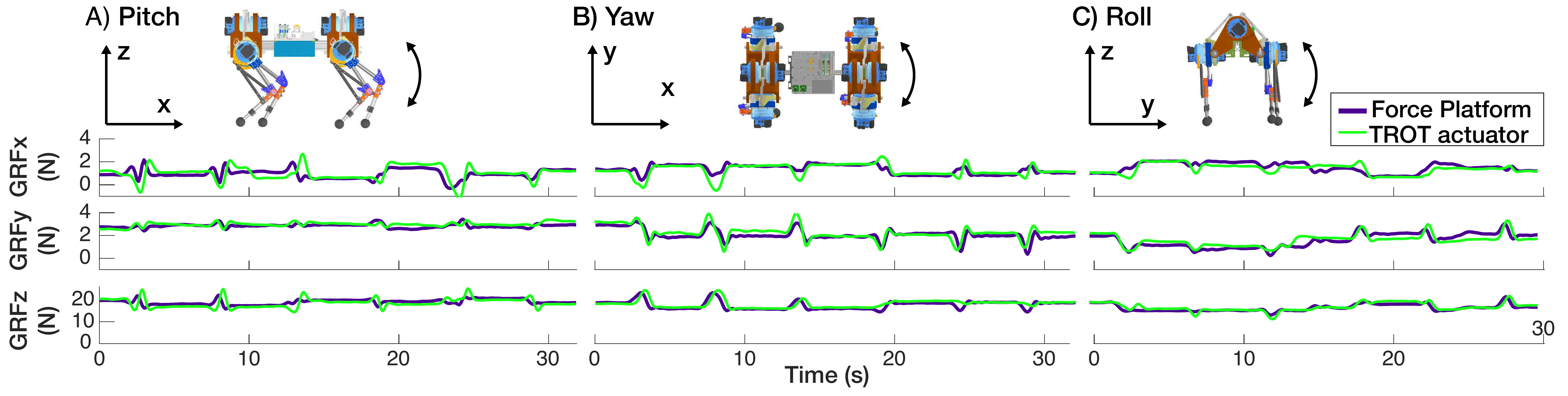}
\caption{
Empirical results for onboard sensor validation for the front right limb during TROT torso rotations in a) pitch, b) roll, and c) yaw.
TROT estimates of ground reaction force for one limb from the proprioceptive actuators (green) were compared to data from a force platform (purple) as the ground truth. 
Six rotations in pitch, yaw, and roll were performed sequentially for each limb.
}
\label{fig:sensor_validation}
\end{figure}

Lastly, by summing the power used by all 12 actuators, we computed the Cost of Transport, $COT$, by dividing by the total robot mass, $M$, gravity, $G$, and velocity, $V$, at which the robot was traveling.
We computed the robot velocity by taking the vectorized Euclidean norm of the velocities in the X, Y, and Z directions resulting from integrating the accelerations recorded by the onboard IMU.

\begin{equation}
COT = \frac{\sum_{n=1}^{12} P }{MGV} 
\label{eq:cot}
\end{equation}

%% file: sections/4results.tex
\section{Results}\label{sec:results}

\subsection{Onboard sensor validation}\label{sec:res:validation}

We performed the same six rotations on TROT four times, changing which foot was on the force platform each time (Fig. \ref{fig:sensor_validation}).
The overall error between the actuator estimates of ground reaction forces and the force platform measurements was relatively high due to the three serial link simplification of the 4-bar linkage (Table \ref{tab:RMSE}).
For a more accurate estimate, the 4-bar linkage version of the URDF can be implemented.

\begin{table}[]
    \centering
    \caption{The root mean squared error between the GRF estimates from the TROT actuators and the force platform in the x, y, and z directions in N.}
    \begin{tabular}{c|cccc}
                  Leg & 3-link X & 3-link Y & 3-link Z \\
                   \hline
        Back Right  & 1.47 & 2.51 &  6.46 \\
        Back Left   & 2.86 & 1.44 &  5.49 \\
        Front Right & 1.83 & 2.83 & 11.15 \\
        Front Left  & 1.29 & 1.67 &  6.84 \\
    \end{tabular}
    \label{tab:RMSE}
\end{table}


\subsection{Effect of limb configuration on limb kinematics}
\label{sec:res:legconfig}

\begin{figure}[ht]
  \centering
\includegraphics[width=1\textwidth]{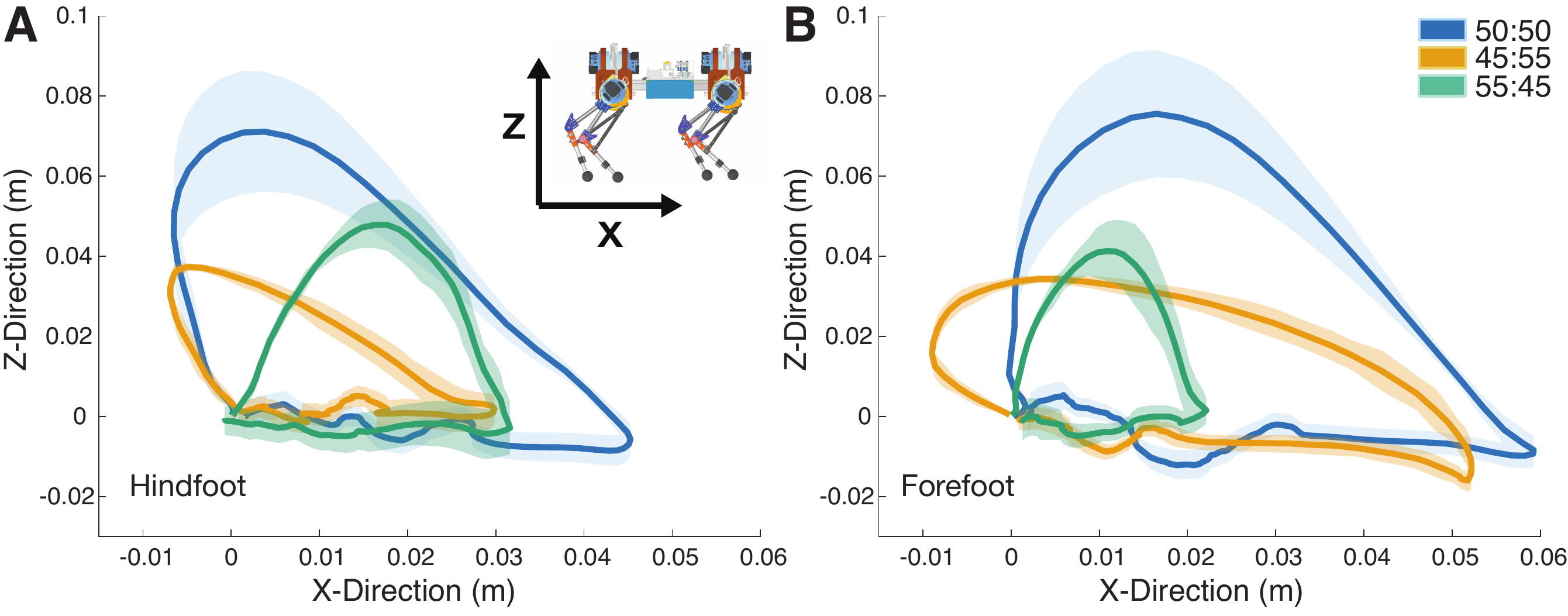}
\caption{Comparisons of TROT (a) hind and (b) fore foot trajectory across different limb length ratios.
The zero value for the X-axis was determined by the hip (actuators X2, X3) center of rotation.
The zero value for the y-axis was the elevation when the foot was in contact with the substrate.
The tracking marker was placed approximately 3.5 cm above the center of the robot foot, along the robot leg, which allows the x-value to fall below zero.
}
\label{fig:foot_traj}
\end{figure}

Changing the limb element length ratio had a dramatic effect on the kinematic trajectory of the end effector, or foot (Fig. \ref{fig:foot_traj}). 
When the tibia and femur were set to  50:50 ratio, the foot trajectory exhibited a pronounced asymmetrical arc in both the forelimb and hindlimb.
Among the configurations tested, the 50:50 leg maximizes the vertical foot displacement and the body translocation per stride.

The 45:55 ratio resulted in decreased foot clearance, with the foot maintaining a relatively even distance from the treadmill surface.
This ratio exhibited the least elevation and the most asymmetry in the shape of the trajectories exhibited by the forefoot and the hindfoot.
Both moved posteriorly (negative X direction) during the beginning of the swing phase before moving anteriorly, but the hindfoot's extreme posterior location was also its extreme elevated position in the swing trajectory.
In contrast, the forefoot extreme elevation was directly below the hip actuators and occurred after the extreme posterior position in swing phase.

Finally, the 55:45 ratio limb generated fore- and hindfoot trajectories that were similar in their symmetrical parabolic shape and low foot elevation.
This limb configuration exhibits the least hip translation per stride.

Each limb configuration also exhibits a significant difference in hip translation between the fore- and hindfoot.
This difference may be the consequence of slipping during stance, or having a longer duty factor in the hindfoot.
Such phenomena are likely destabilizing during a trotting gait, in which contralateral foot pairs are synchronized.


\subsection{Effect of limb inertia on Cost of Transport}\label{sec:res:leginertia}

In its unloaded state, TROT had a COT of 126.12 while moving at 1~m/s for 2~s with a trotting gait (Tab. \ref{table:trot_metrics}).


\begin{table}[ht]
    \centering
    \caption{TROT performance while moving at 1~m/s with a trotting gait for 2 seconds.
    Cost of Transport is a dimensionless number.
    Values represent the sum of all motors.}
    \begin{tabular}{c | c|c|c}
    & Unloaded & Femur & Tibia\\
    \hline
    Peak Power (W)      & 9584.20  & 18603.1    & 21109.03\\  
    Mean Power (W)      & 5070.20  & 9425.62    & 9636.00\\ 
    Peak Torque (Nm)    & 21.59    & 25.49      & 27.70\\
    Mean Torque (Nm)    & 10.66    & 13.19      & 13.85\\ 
    Cost of Transport   & 126.12   & 135.57     & 195.41\\
    \end{tabular}
\label{table:trot_metrics}
\end{table}

Placing 500~g weights on the femora of TROT's limbs resulted in each limb having a moment of inertia of $0.062$~kgm$^2$ about the hip.
The femur-loaded TROT COT was 135.57, a 7.5\% increase in comparison to the unloaded state.

Placing the same weight on the tibiae resulted in a limb moment of inertia of $0.080$~kgm$^2$, a 29\% increase compared to the femur-mounted weighted condition.
The tibia-loaded TROT COT was 195.41, a 44.1\% increase in COT from femur to tibia-loaded conditions, and a 54.94\% increase from unloaded to tibia-loaded conditions.

%% file: sections/newdiscussion.tex
\section{Discussion}\label{sec:disc}
The Robot of Theseus (TROT) is a morphologically modular, shape-shifting quadrupedal robot designed specifically to facilitate hypothesis testing and rapid prototyping of robot designs. 
Unlike most commercially available quadrupedal platforms, such as the MIT Mini Cheetah, Ghost Robotics' Minitaur, or Boston Dynamics' SPOT \cite{minicheetah2019, kenneally2016design, spot}, TROT allows researchers to alter limb structures and configurations in addition to modifying gait patterns within fixed morphologies.
This modularity is particularly advantageous for empirically evaluating hypotheses about limb-specific morphologies and their influence on locomotor performance, bridging gaps often left by purely computational or theoretical approaches.
For example, a previous study investigated the effects of changes in limb morphology on transient quadrupedal locomotion dynamics, but the study was only performed in simulation because a robot capable of these morphological changes did not yet exist \cite{Raw2019}.
Another study introduces a modular quadrupedal robot, but the servo motors used to actuate the legs cannot achieve the necessary forces and speeds to mimic the leg kinematics of many animals \cite{geva2014novel}.
By combining telescoping and modular legs with high-torque, high-speed, low mass, proprioceptive actuators, TROT directly addresses limitations of conventional quadrupedal robots by enabling physical experiments previously confined to simulations.
Such studies can greatly inform the design of future legged robotic limbs \cite{mckenzie2012design}.

TROT also has immense potential to inform biological studies, due to its limb morphologies mapping to a broad swathe of biologically relevant limb proportions.
Much paleontological research considers the effect of evolutionary changes in limb morphology on locomotion performance, such as peak velocity, cost of transport, and effectiveness on different terrains \cite{SmithSavage1956, Janis2020, fischer2002basic, polly2007}.
These studies are informed by research on live animals \cite{Biewener1983a}, but morphologies found in extinct animals often occupy gaps in the extant animal morphospace.
Furthermore, comparisons between animals that have variation in the parameter of choice can be muddled by evolutionary variation in morphology and physiology that are not being directly studied \cite{taylor1974running}.
These studies highlight the need for versatile experimental platforms that can be precisely varied to examine the biomechanical consequences of evolutionary changes in morphology.
As a robotic quadruped designed with shape-shifting limb morphologies, we expect that TROT will be as informative to the field of locomotion functional morphology as other robotic physical models have been to understanding the biomechanical evolution of swimming and flying \cite{flyingfish2022, lauder2007fish,koehl2011using,peterson2011wing}.
Already, our findings demonstrate that even small modifications in leg morphology result in significant changes in foot kinematics, underscoring the sensitivity of locomotion dynamics to morphological variations.


A key innovation of TROT is its capability to estimate ground reaction forces (GRFs) directly through its leg actuators. 
The vast majority of quadrupedal legged robots include series-elastic elements in their legs, making it difficult to use their QDD actuators accurately estimate the external load on the actuators \cite{murphy2011littledog,hutter2016anymal}.
Therefore, these robots are generally run on bulky and flat force-platforms, making it difficult to study and optimize the dynamics of locomotion on natural terrain \cite{lee2021whole}.
To measure ground-reaction forces in uncontrolled, outdoor conditions, such robots would require force transducers on the tip of each foot \cite{valsecchi2020quadrupedal,ruppert2020foottile}, a placement that subjects the sensor to repeated collisions that can cause damage.
Instead, we leveraged the tunable virtual stiffness and backdrivability of QDD actuators to maintain the spring-like behavior of the leg, while removing all series elastic elements to more accurately estimate ground-reaction force by measuring the external loads on the actuators \cite{urs2022design,blackman2016gait}.
This feature dramatically expands the robot's potential for hypothesis testing, enabling detailed locomotion dynamics studies on various natural and uneven substrates without specialized infrastructure.

Overall, we were motivated to construct a quadrupedal robot capable of sophisticated locomotion control to bridge the gap between biology and robotic locomotion research.
Commercially available robots are too expensive for most biologists to justify, the ability to alter their morphology is limited, and adjusting the controller beyond the user-interface requires specialized training in the robot's programming language.
Instead, TROT's design is intentionally cost-effective and accessible, facilitating widespread use and rapid iteration in both biological research and robot design testing. 
This approach not only democratizes experimental capabilities but also significantly reduces the barrier between simulation and real-world testing, helping to mitigate the sim-to-real gap prevalent in robotics. 
Consequently, TROT represents a useful robo-physical platform, balancing biological relevance and experimental generalizability, suitable for exploring both existing biological morphologies and theoretical or extinct limb structures.